\RequirePackage[T1]{fontenc}
\documentclass[letterpaper,10pt,conference]{ieeeconf}

\IEEEoverridecommandlockouts
\usepackage{amsmath,amssymb}
\usepackage{booktabs}
\usepackage{multirow}
\usepackage{graphicx}
\usepackage{xcolor}
\usepackage{url}
\urldef\codeurl\url{https://github.com/MINT-SJTU/Evo-RL}
\urldef\dataseturl\url{https://huggingface.co/datasets/MINT-SJTU/RW-RL-Dataset}
\usepackage{algorithm}
\usepackage{algpseudocode}

\algrenewcommand{\algorithmicrequire}{\textbf{Input:}}
\algrenewcommand{\algorithmicensure}{\textbf{Output:}}

\title{ \Large \bf
Evo-RL: Iterative Robot Reinforcement Learning from Selective Experience
}

\author{\textcolor{black}{Kai Ye$^{1}$, Yifan Jin$^{1,2,\dagger}$, Qianhang Luo$^{1,3,\dagger}$, Hongru Zhang$^{1,\dagger}$, Jiedong Zhang$^{1}$,}\\
\textcolor{black}{Shuyuan Hu$^{1}$, Song-Yuan Wu$^{4}$, Jihai Xu$^{4}$, Jie Zhao$^{4}$, Bo Zhao$^{1,*}$}%
\thanks{\textcolor{black}{$^{\dagger}$Equal Contribution. $^{*}$Corresponding author.}}%
\thanks{\textcolor{black}{Correspondence: \texttt{bo.zhao@sjtu.edu.cn}.}}%
\thanks{\textcolor{black}{$^{1}$School of AI, Shanghai Jiao Tong University. $^{2}$Central South University. $^{3}$South China Normal University. 
$^{4}$Boden AI.}}%
\thanks{\textcolor{black}{Code: \codeurl}}%
\thanks{\textcolor{black}{Dataset: \dataseturl}}%
}

\begin{document}
\bstctlcite{rwrl_bib_style}
\maketitle
\thispagestyle{empty}
\pagestyle{empty}

\begin{abstract}
\textcolor{black}{Existing real-world reinforcement learning methods improve robot policies by combining human corrections with autonomous rollouts. However, accumulated experience varies in its value for policy improvement, and even successful trajectories can contain inefficient or redundant behavior, motivating more selective experience reuse. To this end, we propose Evo-RL, an iterative reinforcement learning method that combines human interventions with advantage-guided experience selection. Evo-RL uses human corrections and experience with positive estimated advantages from successful trajectories to iteratively refine the policy. To support learning from these different experience sources, we introduce RW-RL-Dataset, which spans over 1,000 hours across four robot embodiments. On a real-world high-precision manipulation task, Evo-RL improves attempt success rate over RECAP by 30.46\%. Our code, model checkpoints, and dataset are publicly available.}
\end{abstract}

\begin{figure*}[t]
\centering
\includegraphics[width=\textwidth]{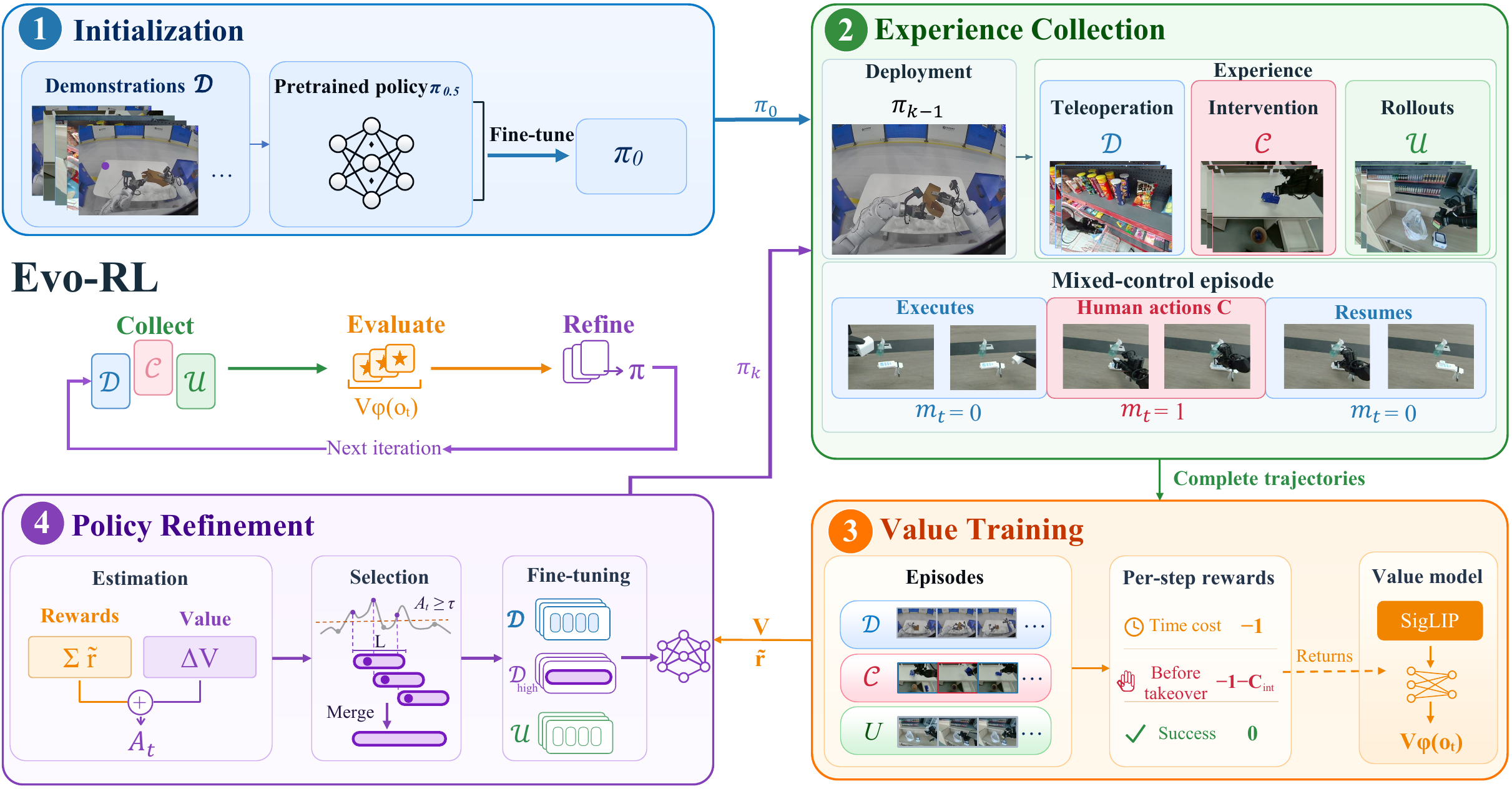}
\caption{\textbf{Overview of the Evo-RL method. }The current policy is deployed to collect demonstrations, human interventions, and autonomous rollouts. Complete trajectories are used to train a value model whose estimates guide policy refinement. The updated policy is then redeployed in the next iteration. }
\label{fig:system_overview}
\end{figure*}

\section{Introduction}
\textcolor{black}{Vision-language-action (VLA) models generate robot actions by conditioning on visual observations and language instructions~\cite{RT-2}. Recent advances have extended their applicability to a wider range of tasks and robot embodiments~\cite{OpenVLA,pi_0,Lingbot} and improved deployment efficiency~\cite{Evo-1}. Yet policies trained on demonstrations can still encounter unfamiliar states during deployment, where errors can accumulate and recovery may be difficult~\cite{ross2011dagger}. These limitations motivate continued policy improvement using experience collected during execution.}

Existing approaches support this improvement through human corrections and reinforcement learning. Human interventions supply corrective behavior at states visited by the deployed policy~\cite{kelly2019hgdagger,mandlekar2020iwr}, while autonomous rollouts provide feedback beyond the initial demonstrations. RECAP~\cite{RECAP} combines demonstrations, autonomous rollouts, and human interventions, using value-estimated advantages to condition policy training and iteratively update the policy with deployment data.

The value of accumulated experience for policy improvement varies across and within trajectories. Even successful trajectories may contain hesitation, repeated attempts, or other inefficient behavior, so the contribution of experience to policy improvement must be assessed at a finer level. Human corrections target deployment difficulties, but policy actions in the same trajectories may include behavior that prompted the intervention. These observations motivate Evo-RL’s selective update procedure: estimated advantages guide the selection of experience for policy updates, while only human-controlled portions of successful intervention trajectories are retained for corrective fine-tuning.

We propose Evo-RL, an iterative real-world reinforcement learning method that combines human corrections with advantage-guided experience selection. Starting from a demonstration-trained policy, Evo-RL collects human interventions and autonomous rollouts during deployment. It trains a value model on demonstrations, intervention data, and autonomous experience, using rewards that account for execution time and human takeovers, and then estimates advantages to guide experience selection. For policy fine-tuning, Evo-RL retains only the human-controlled portions of successful intervention trajectories, keeps the original demonstrations, and adds selected positive-advantage experience from demonstrations and successful autonomous rollouts. RECAP trains the policy on accumulated experience with a binary advantage label as an additional input. Evo-RL instead uses estimated advantages to select experience for policy updates and increase the sampling weight of selected experience. The updated policy is then redeployed for the next iteration.

\textcolor{black}{To advance the real-world RL research, we further introduce RW-RL-Dataset which organizes demonstrations, human interventions, and autonomous experience with action-source metadata and semantic annotations. It spans over 1,000 hours across four diverse robot embodiments and includes the task-specific training experience used in our experiments.}

Our main contributions are as follows:
\begin{itemize}
    \item \textcolor{black}{We propose Evo-RL, an iterative method that combines corrective supervision with selective reuse of experience guided by positive estimated advantages. Under our evaluation protocol, first-round Evo-RL achieves an attempt success rate (ASR) 30.46\% higher than RECAP on a real-world manipulation task requiring high-precision alignment.}
    \item \textcolor{black}{We present RW-RL-Dataset, a collection of over 1,000 hours of real-world data from four robot embodiments, pairing synchronized camera observations with robot states and actions. Its annotations specify task instructions and scene context, execution quality and error counts, subtask intervals, and whether each frame records human or policy control.}
\end{itemize}

\section{Related Work}
\subsection{Vision-Language-Action Model and Robot Datasets}
Large-scale robot datasets collectively cover diverse household, office, and industrial manipulation tasks across single-arm, dual-arm, and humanoid platforms~\cite{walke2023bridgedatav2,khazatsky2024droid,fang2023rh20tcomprehensiveroboticdataset,wu2024robomind,openx2023}. This diversity provides a foundation for VLA pretraining, enabling policies to learn reusable manipulation skills across tasks and robots~\cite{openx2023}. Building on these skills, models such as OpenVLA can adapt to new tasks and robot setups through fine-tuning on additional demonstrations~\cite{OpenVLA}. However, even after fine-tuning, policies may still fail in situations poorly covered by datasets. Human interventions provide targeted corrections in these situations, yet the resulting corrective experience is less systematically captured and organized.

\begin{figure*}[t]
\centering
\includegraphics[width=\textwidth]{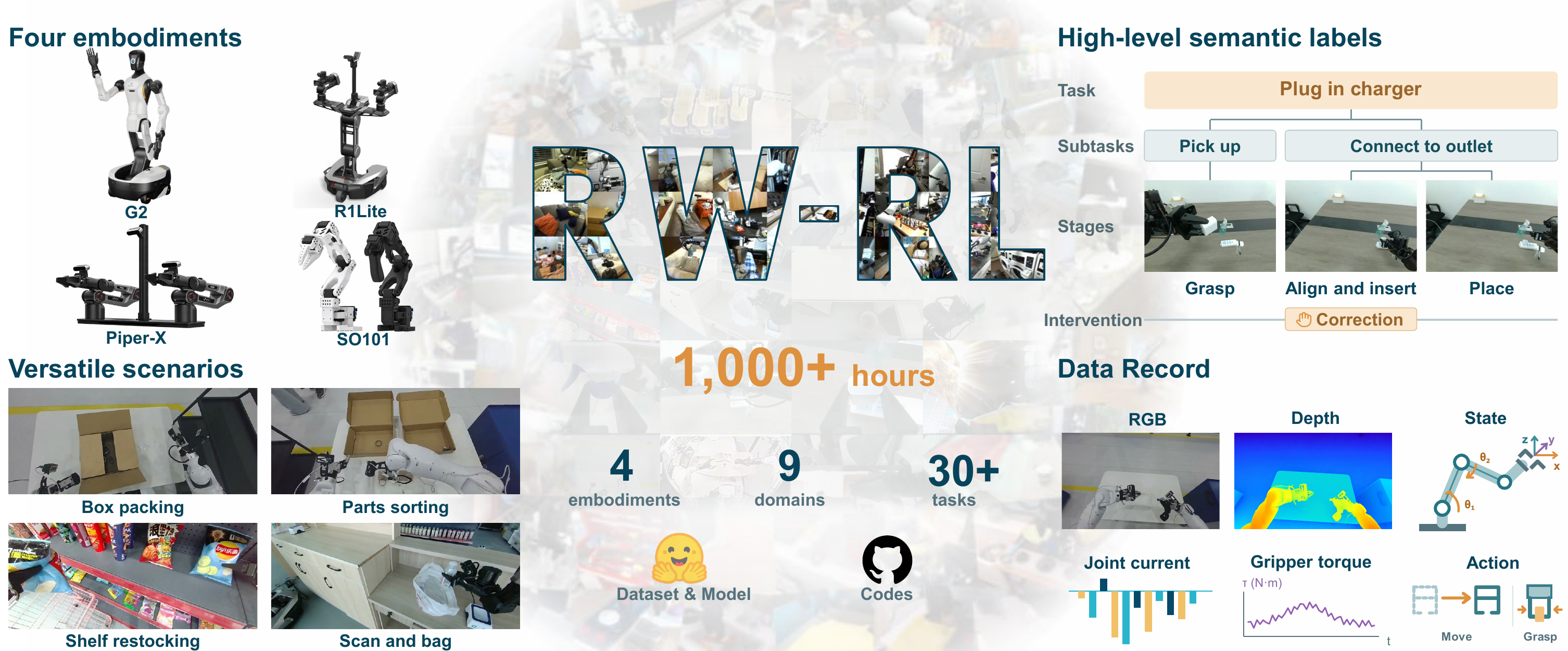}
\caption{Overview of RW-RL-Dataset, a large-scale real-world robot learning dataset spanning diverse embodiments and tasks, with hierarchical semantic annotations, synchronized multimodal observations and control signals, and over 1,000 hours of real-world interaction data.}
\label{fig:data_overview}
\end{figure*}

\subsection{Human Interventions and Data Aggregation}
Out-of-distribution (OOD) situations refer to scenarios that differ from those encountered during training and may therefore challenge the policy’s ability to generalize. During deployment, human interventions can be introduced in such OOD situations, not necessarily only after an execution error occurs. These interventions provide corrective demonstrations that can be incorporated into the training data, helping the policy improve its responses when similar situations arise in the future. Dataset Aggregation (DAgger) formalizes this process by querying expert actions for states visited by the current policy and aggregating the resulting state--action pairs into the training set~\cite{ross2011dagger}.

Building on this paradigm, subsequent research focuses on two related questions: when human interventions should be triggered and how the resulting correction data should be incorporated into policy learning. Some methods let human operators decide when to intervene~\cite{kelly2019hgdagger}, whereas others use state novelty or failure risk to determine when to request human assistance~\cite{hoque2022thriftydagger}. Because intervention samples often constitute only a small fraction of the collected data, their influence on policy updates can be limited. IWR addresses this imbalance by increasing the relative weight of intervention samples, while using non-intervention samples to regularize policy updates and help retain previously learned skills~\cite{mandlekar2020iwr}. However, corrective action labels alone do not explicitly indicate which states are likely to lead to failure or which trajectories are of higher quality. Incorporating such evaluations can provide additional supervisory signals for further policy improvement.

\subsection{Real-World RL and Iterative Policy Improvement}
Reinforcement learning enables robots to improve their policies through interaction with the environment. Early work on RL for VLA models focused on offline preference alignment\cite{zhang2025reinbotamplifyingrobotvisuallanguage,zhang2025grapegeneralizingrobotpolicy}, such as GRAPE \cite{zhang2025grapegeneralizingrobotpolicy}, which employs Direct Preference Optimization (DPO) \cite{rafailov2024directpreferenceoptimizationlanguage} to align robot trajectories with human intent. More recent approaches incorporate experience collected during deployment into VLA post-training. RECAP trains a value function on demonstrations, human interventions and autonomous rollouts then uses advantage-based indicators to condition policy training and guide iterative improvement\cite{RECAP}. To reduce the cost of online adaptation, RLT compresses latent representations into an RL token and trains a lightweight actor--critic to refine action chunks while keeping the VLA frozen\cite{RLT}. Beyond action-level optimization, LaST-R1 extends reinforcement learning to the latent space, jointly optimizing reasoning and execution\cite{LaST-R1}. Our Evo-RL is motivated by RECAP, as both methods combine human corrections with autonomous rollouts to realize value-guided iterative policy improvement.

\section{Method}
\label{sec:method}
Evo-RL organizes policy improvement into a data loop with demonstration-based initialization followed by two complementary stages: fine-tuning with human interventions and advantage-guided policy refinement. Starting from demonstration data, we initialize the policy by fine-tuning a pretrained VLA model. Next, human interventions collected during deployment are used for policy fine-tuning to improve performance in challenging scenarios. Following intervention-based refinement, we use accumulated experience for advantage-guided policy updates. Algorithm~\ref{alg:evo_rl} summarizes this loop and the policy updates within each iteration.

\begin{algorithm}[t]
\caption{Evo-RL: Iterative \textcolor{black}{Robot Reinforcement Learning} with Selective Experience}
\label{alg:evo_rl}
\small
\begin{algorithmic}[1]

\Require Policy $\pi_K$,
demonstrations $\mathcal D_{\mathrm{demo}}$,
human corrections $\mathcal D_{\mathrm{interv}}$,
rounds $R$.
\Ensure Final policy $\pi_{K+1+2R}$.

\State \textbf{Demonstration-Based Initialization}
\State $\pi_{K+1}\leftarrow
\operatorname{SFT}(\pi_K,\mathcal D_{\mathrm{demo}})$.

\State Initialize
$\mathcal D_{\mathrm{auto}}$,
$\mathcal D'_{\mathrm{auto}}$,
$\mathcal D'_{\mathrm{demo}}$, and
$\mathcal D_{\mathrm{train}}$ as empty sets.

\For{$k=K+1,K+3,\ldots,K+2R-1$}

    \State \textbf{1. Policy Refinement with Human Interventions}
    \State Deploy $\pi_k$ to collect human corrections
    and update $\mathcal D_{\mathrm{interv}}$.
    \State $\pi_{k+1}\leftarrow
    \operatorname{SFT}\bigl(
        \pi_k,
        \operatorname{Mix}(
            \mathcal D_{\mathrm{demo}},
            \mathcal D_{\mathrm{interv}})
    \bigr)$.

    \State \textbf{2. Advantage-Guided Policy Fine-Tuning}
    \State Optionally deploy $\pi_{k+1}$ to collect autonomous
    rollouts and update $\mathcal D_{\mathrm{auto}}$.
    \State Train $V_\phi$ on
    $\mathcal D_{\mathrm{demo}}$, 
    $\mathcal D_{\mathrm{interv}}$ and $\mathcal D_{\mathrm{auto}}$
    \State Compute $\widehat A_t$ on
    $\mathcal D_{\mathrm{demo}}$ and
    $\mathcal D_{\mathrm{auto}}$.
    \State Select high-advantage segments
    as $\mathcal D'_{\mathrm{demo}}$
    and $\mathcal D'_{\mathrm{auto}}$.
    \State $\mathcal D_{\mathrm{train}}\leftarrow
    \operatorname{Mix}\bigl(
        \mathcal D_{\mathrm{demo}},
        \mathcal D'_{\mathrm{auto}},
        \mathcal D_{\mathrm{interv}}
    \bigr)$.
    \State $\pi_{k+2}\leftarrow
    \operatorname{SFT}(\pi_{k+1},\mathcal D_{\mathrm{train}})$.

\EndFor

\Statex \Comment{Segments in
$\mathcal D'_{\mathrm{demo}}\subseteq\mathcal D_{\mathrm{demo}}$
are sampled with elevated probability.}

\end{algorithmic}
\end{algorithm}
\subsection{Demonstration-Based Policy Initialization}
At time step $t$, the policy receives an observation $o_t = (I_t, s_t)$ and a task instruction $\ell$, where $I_t$ denotes the camera images and $s_t$ the robot's proprioceptive state. The policy models a conditional distribution over action chunks:
\begin{equation}
    \mathbf{a}_t = (a_t, \ldots, a_{t+H-1})
    \sim \pi_\theta(\cdot \mid o_t, \ell),
\end{equation}
where $H$ is the chunk length and $\theta$ denotes the policy
parameters.
Following the architecture of $\pi_{0.5}$~\cite{black2025pi05}, our policy combines a PaliGemma vision-language backbone with a flow-matching action expert to generate action chunks conditioned on visual observations, proprioceptive state, and task instructions. We train the policy on teleoperated demonstrations $D_{demo}$ using the flow-matching objective.

For each demonstration sample $(o_t,\ell,\mathbf{a}_t)\sim D_{demo}$,
we sample standard Gaussian noise $\epsilon$ with the same
shape as $\mathbf{a}_t$, and a flow time $\tau\in[0,1]$.
The noisy action chunk is constructed as
\begin{equation}
    x_t^\tau = (1-\tau)\mathbf{a}_t + \tau\epsilon.
\end{equation}
The action expert $v_\theta$ predicts the corresponding
velocity $\epsilon-\mathbf{a}_t$, conditioned on $o_t$
and $\ell$. The objective function can be written as
\begin{equation}
    \mathcal{L}_{\mathrm{demo}}(\theta)
    =
    \mathbb{E}\!\left[
        \operatorname{MSE}\!\left(
            v_\theta(x_t^\tau,o_t,\ell,\tau),
            \epsilon-\mathbf{a}_t
        \right)
    \right],
\end{equation}

While this objective enables effective behavioral cloning, the model’s capability and generalization remain inherently bounded by the quality, scale, and diversity of the demonstration data. This fine-tuning produces the initial task policy $\pi_0$,
which is subsequently deployed to collect experience for further improvement.

\begin{figure*}[t]
\centering
\includegraphics[width=\textwidth]{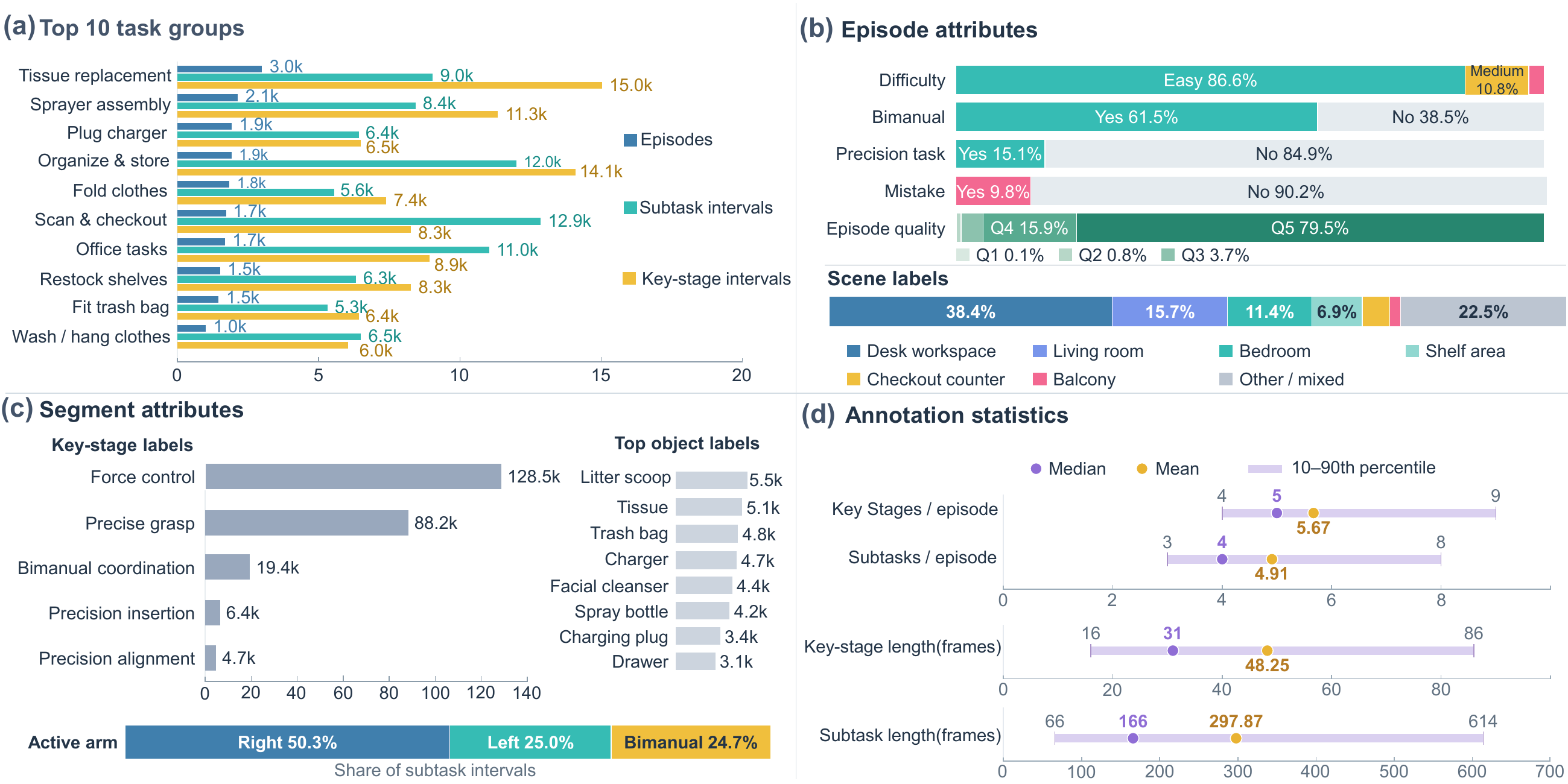}
\caption{Annotation statistics of RW-RL-Dataset, summarizing task coverage, episode-level attributes, segment-level labels, and the counts and durations of annotated subtasks and key stages.}
\label{fig:data_characterization}
\end{figure*}

\subsection{Policy Refinement with Human Interventions}
The initial policy may still struggle in challenging scenarios. We therefore use human-supervised rollouts to collect corrective actions. An operator monitors task execution and intervenes when the policy encounters states poorly covered by the demonstrations and struggles to recover or make further progress. After each correction, the robot resumes autonomous operation.
For an intervention trajectory indexed by $i$, we represent the information collected at each frame $t$ as
\[ S_{i,t} = (\ell_i, o_{i,t}, a_{i,t}, m_{i,t}), \]
where $\ell_i$ is the task instruction, $o_{i,t}$ contains images and proprioceptive state, and $a_{i,t}$ is the executed action. We introduce the control-source label $m_{i,t}$ to distinguish human corrections from policy actions, with $1$ indicating human control and $0$ indicating policy control. A trajectory of length $T_i$ is written as
\[ \xi_i = (S_{i,1}, \ldots, S_{i,T_i}), \]
with $y_i \in {0,1}$ indicating whether the task succeeds. We retain human-controlled data from successful intervention trajectories to construct the dataset $D_{\mathrm{interv}}$:
\[ D_{\mathrm{interv}} = \left\{ S_{i,t} \mid y_i = 1,\ m_{i,t} = 1 \right\}. \]
We preserve the temporal order and segment boundaries of these data and sample training action chunks only within contiguous human-controlled segments. Policy-controlled portions are excluded from $D_{\mathrm{interv}}$ to avoid learning from actions that are likely to cause failures or prolong task completion.
During training, we fine-tune the current policy with the same loss function, sampling from $D_{\mathrm{interv}}$ and $D_{demo}$ with probabilities $p \in (0,1)$ and $1-p$, respectively.
The updated policy is then redeployed to identify remaining difficulties and collect further corrections. This cycle continually incorporates human corrections into training, allowing the policy to learn from its mistakes and improve task performance over successive updates.

\subsection{Reinforcement Learning for Policy Refinement} \textcolor{black}{Demonstration} training and intervention-based refinement \textcolor{black}{provide an initial policy} and \textcolor{black}{corrective supervision. Further improvement requires} finer-grained signals beyond episode-level success or failure to identify \textcolor{black}{promising} action segments \textcolor{black}{within the accumulated experience. We use rewards that account for execution time and human takeovers} to \textcolor{black}{learn a value model and} estimate segment \textcolor{black}{advantages. These estimates guide experience selection and sampling} for \textcolor{black}{policy fine-tuning in each round.}

\subsubsection{Reward Design with Time and Intervention Costs}

To account for both execution time and reliance on human corrections, we construct rewards from per-frame costs and
takeover events. Using the label $m_t$ introduced in the previous subsection, we identify a takeover as
\begin{equation}
    b_{t+1} = \mathbf{1}[m_t = 0,\ m_{t+1} = 1].
\end{equation}
For a successful trajectory, let $T$ denote the index of the terminal observation and set $b_T = 0$.
The frame-level reward is
\begin{equation}
    r_t =
    \begin{cases}
        0, & t = T, \\
        -1 - C_{\mathrm{int}}b_{t+1}, & t < T,
    \end{cases}
    \label{eq:frame_reward}
\end{equation}
where $C_{\mathrm{int}} \geq 0$ is the additional cost of a human takeover.

A takeover typically indicates that the policy's behavior requires correction. We therefore treat it as negative feedback
and assign the penalty to the policy transition immediately preceding the takeover, rather than to the subsequent human
corrections~\cite{luo2024rlif}. The unit cost at each nonterminal frame encourages faster task completion. Together, these terms favor trajectories with lower execution time and less dependence on human assistance.
\subsubsection{Value Function Learning from Trajectory Returns}

For each task, we train a vision-only value model with parameters
$\phi$ independently of the policy. Given camera images $I_t$,
it predicts a categorical distribution $p_\phi(c \mid I_t)$
over 201 bins whose centers $v_c$ are evenly spaced in $[-1,0]$.

The return of frame $t$ is defined as $G_t = \sum_{u=t}^{T} r_u$. We normalize it by a fixed,
task-specific scale $Z_\ell > 0$, giving a training target $G_t/Z_\ell \in [-1,0]$. Returns closer to zero indicate a
lower combined cost of remaining execution time and human takeovers.

Each normalized return is encoded as a two-hot target $q_t(c)$,
with weights determined by linear interpolation between adjacent
bin centers. We train the model on sampled trajectory frames
using the cross-entropy loss
\begin{equation}
    \mathcal{L}_V(\phi)
    =
    -\mathbb{E}_{\xi,t}
    \left[
        \sum_{c=1}^{201}
        q_t(c)\log p_\phi(c \mid I_t)
    \right].
    \label{eq:value_loss}
\end{equation}
The scalar value estimate is the probability-weighted mean
of the bin centers:
\begin{equation}
    V_\phi(I_t)
    =
    \sum_{c=1}^{201}
    p_\phi(c \mid I_t)v_c.
\end{equation}

\subsubsection{Advantage-Guided Policy Fine-Tuning}
 
Using the trained value model, we estimate the advantage of an action sequence as
\begin{equation}
    \widehat{A}_t
    =
    \sum_{j=0}^{n_t-1}\tilde{r}_{t+j}
    +
    V_\phi(o_{t+n_t},\ell)
    -
    V_\phi(o_t,\ell).
    \label{eq:advantage}
\end{equation}
Here, $\tilde{r}_t = r_t/Z_\ell$ is the normalized reward,
and $n_t = \min(n,T-t)$ limits the lookahead window of at
most $n \leq H$ frames to the remaining trajectory.
The terminal value is fixed to $V_\phi(o_T,\ell)=0$.

In each training round, we use estimated advantages to select promising segments from $D_{demo}$ and available $D_{auto}$. These segments are mixed with human corrections to fine-tune the policy with the same flow-matching loss. Selection and sampling details are given in Section~\ref{sec:advantage_selection}.

\section{RW-RL-Dataset}
\label{sec:dataset}
RW-RL-Dataset organizes demonstrations, human interventions, and autonomous executions to support research on iterative robot policy improvement.
Figure~\ref{fig:data_overview} summarizes four key dimensions of RW-RL-Dataset: robot embodiment diversity, long-horizon real-world executions, frame-aligned semantic annotations, and multi-source collection regimes for iterative policy learning.

\begin{figure*}[t]
\centering
\includegraphics[width=\textwidth,trim=0 223bp 0 0,clip]{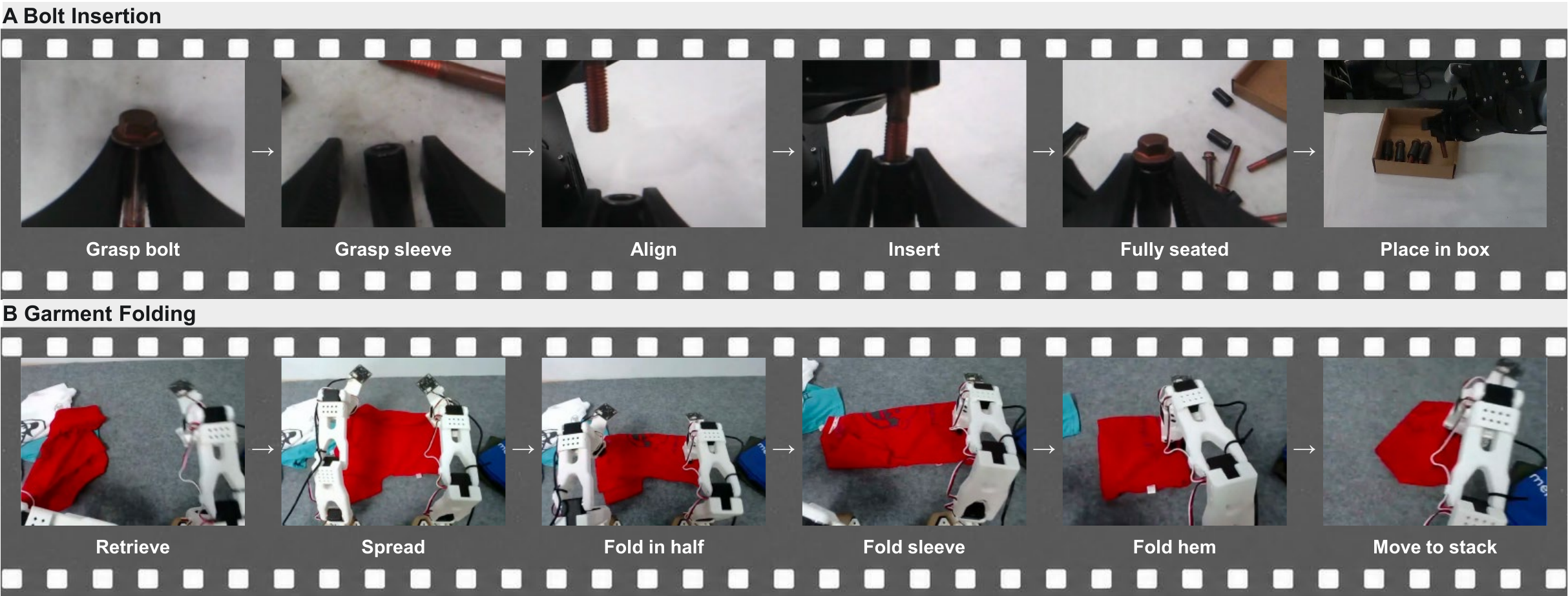}
\caption{Task sequences in RW-RL-Dataset: (A) bolt insertion and (B) garment folding. Each row shows one episode in chronological order from left to right.}
\label{fig:task_progression}
\end{figure*}

\subsection{Embodiments, Modalities, and Collection Regimes}

The dataset contains more than 1,000 hours of real-world interaction collected on four robot embodiments: AgiBot G2, Galaxea R1Lite, AgileX PiperX, and SO-101. These platforms span industrial mobile manipulation, modular dual-arm systems, and low-cost tabletop learning.

G2 provides eight RGB video streams (top, top-back, paired fisheye and stereo views, and two wrist views), together with joint states, end-effector poses and wrenches, gripper/head/waist/base states, and actions. R1Lite provides four RGB streams (top, top-right, and two wrist views), together with arm joint positions and velocities, gripper/torso/chassis states, chassis IMU readings, end-effector poses, and actions. The PiperX and SO-101 platforms are dual-arm systems. Both are fitted with an external RGB camera and a pair of wrist cameras. PiperX uses 14-dimensional state and action vectors, while SO-101 employs 12-dimensional counterparts.

Beyond embodiment diversity, RW-RL-Dataset also covers a broad range of real-world task settings. The data span nine scene domains: store shelves, industrial production, commercial kitchens, home spaces, care environments, laboratories, warehouses, hotels, and education or training settings. Across these domains, the dataset instantiates more than 30 task templates, including object sorting, shelf replenishment, scanning and bagging, precision insertion, garment folding, and box stacking.

RW-RL-Dataset further distinguishes episodes by how the actions are generated during collection. Demonstrations record human execution; autonomous episodes record policy execution; and intervention episodes combine both, with frame-level controller labels identifying human correction segments.

\subsection{Data Organization and Hierarchical Semantic Annotations}
Each episode in the dataset links visual observations, robot states, and executed actions through episode identifiers and timestamps. These records are accompanied by semantic annotations at three levels of granularity: dataset, episode, and frame (Fig.~\ref{fig:data_characterization}). 
Dataset-level fields specify the task, embodiment, difficulty, objects, and coordination or precision requirements. 
Episode-level fields describe instructions, scenes, outcomes, execution quality, and mistakes. They also include action-source metadata, which categorizes episodes into three types: teleoperated demonstrations, autonomous policy rollouts, and human-intervention episodes.
Frame-level annotations consist of subtask intervals, along with frame-wise controller labels that identify human and policy actions. A subtask interval records its start and end frames, local instruction, manipulated objects, and active arm, and marks operation attributes such as bimanual coordination. All these annotations are synchronized under a unified timestamp system for straightforward cross-reference.

\section{Experiments}
\label{sec:experiments}
\begin{table*}[t]
\centering
\caption{Targeted interventions improve PiperX ASR under matched active human-control budgets, which include demonstration and intervention time. Each method uses $N=10$ planned instances per unit.}
\label{tab:scale_intervention}
\begin{tabular*}{\textwidth}{@{\extracolsep{\fill}}clccccccc@{}}
\toprule
& & \multicolumn{5}{c}{Successes / attempts} & & \\
Budget & Method & $U_1$ & $U_2$ & $U_3$ & $U_4$ & $U_5$ & ASR (\%) & TP \\
\midrule
\multirow{2}{*}{2 h} & SFT & 0/18 & 0/0 & 0/0 & 0/0 & 0/0 & 0.00 & 0.00 \\
& SFT + Int. & 5/12 & 1/1 & 0/0 & 0/0 & 0/0 & 46.15 & 7.20 \\
\midrule
\multirow{2}{*}{4 h} & SFT & 5/23 & 4/7 & 1/7 & 0/0 & 0/0 & 27.03 & 12.00 \\
% Measured 4 h (3+1) evaluation: 17 successes / 33 attempts over 3000 s; blank units were not started.
& SFT + Int. & 7/13 & 5/14 & 3/3 & 2/2 & 0/1 & 51.52 & 20.40 \\
\midrule
\multirow{2}{*}{8 h} & SFT & 9/21 & 9/14 & 8/23 & 8/19 & 4/10 & 43.68 & 50.17 \\
% Measured 8 h (7+1): 39 successes / 50 attempts over 2647 s; unstarted units contribute zero attempts.
& SFT + Int. & 9/11 & 9/10 & 8/12 & 7/10 & 6/7 & 78.00 & 53.04 \\
\midrule
16 h & SFT & 9/12 & 8/9 & 8/8 & 6/12 & 5/5 & 78.26 & 53.27 \\
\bottomrule
\end{tabular*}
\end{table*}

We first assess whether intervention data are more effective than additional demonstrations under the same data budget
(Table~\ref{tab:scale_intervention}).
Beyond this data-efficiency comparison, task-specific metrics are used to evaluate policy improvements across successive training stages (Tables~\ref{tab:complete_loop}).
Representative sequences from the dataset illustrate the progression of each task in Figure~\ref{fig:task_progression}. We also describe our value-model choice and advantage-guided training-data mixture.
\subsection{Experimental Setup and Task Definitions}
\label{sec:exp_platforms}
We conduct experiments on two robot platforms: PiperX and SO-101. 
Policies are initialized from $\pi_{0.5}$ and deployed with
real-time chunking (RTC)~\cite{black2025rtc}.\par
Each trial has a five-minute time limit. A trial is considered a failure if the task is not completed within this limit. The setups for the two manipulation tasks are described below.\par
PiperX insertion. Five screw--sleeve pairs are placed on the tabletop. The robot selects, grasps, aligns, and inserts each screw. Either arm can insert the screw into the sleeve held by the other.  Each unit $U_j$ corresponds to inserting one screw into a sleeve.\par
SO-101 folding. Five T-shirts are loosely piled on the desk. The robot retrieves and folds them sequentially. Units $U_1$--$U_5$ denote retrieval, spreading, folding in half, folding the sleeves, and folding the hem toward the collar. If the garment enters an erroneous state, an operator manually restores it to the state before the failed stage without fully smoothing it out.\par
\begin{table*}[t]
\centering
\caption{Combining corrections with RL improves PiperX ASR and throughput beyond either component alone; a second round yields further gains. Each method uses $N=10$ planned instances per PiperX unit and $N=20$ per SO-101 folding stage.}
\label{tab:complete_loop}
\begin{tabular*}{\textwidth}{@{\extracolsep{\fill}}clccccccc@{}}
\toprule
& & \multicolumn{5}{c}{Successes / attempts} & & \\
Task & Method & $U_1$ & $U_2$ & $U_3$ & $U_4$ & $U_5$ & ASR (\%) & TP \\
\midrule
\multirow{2}{*}{SO-101 folding} & SFT & 20/20 & 17/76 & 16/18 & 16/21 & 16/18 & 55.56 & 75.78 \\
& Evo-RL & \textcolor{black}{20/20} & \textcolor{black}{19/36} & \textcolor{black}{18/22} & \textcolor{black}{18/22} & \textcolor{black}{17/20} & \textcolor{black}{76.67} & \textcolor{black}{90.00} \\
\midrule
\multirow{6}{*}{PiperX insertion} & SFT & 7/20 & 3/15 & 0/9 & 0/0 & 0/0 & 22.73 & 12.00 \\
& SFT + Int. & 9/11 & 9/12 & 8/10 & 6/8 & 5/6 & 78.72 & 47.98 \\
% Latest user-supplied 10-record SFT+RL subset: 27 successes / 73 attempts over 2885 s.
% This subset excludes the earlier 11-record set's U1 failure (4 attempts, 300 s); exclusion rationale remains undocumented.
& SFT + RL & 10/25 & 8/21 & 5/14 & 3/10 & 1/3 & 36.99 & 33.69 \\
& \textcolor{black}{RECAP} & \textcolor{black}{8/13} & \textcolor{black}{6/10} & \textcolor{black}{5/11} & \textcolor{black}{4/5} & \textcolor{black}{3/6} & \textcolor{black}{57.78} & \textcolor{black}{35.54} \\
& Evo-RL & 10/10 & 10/10 & 10/12 & 10/11 & 5/8 & 88.24 & 60.61 \\
& Evo-RL (R2) & \textcolor{black}{10/10} & \textcolor{black}{10/10} & \textcolor{black}{10/11} & \textcolor{black}{9/10} & \textcolor{black}{8/10} & \textcolor{black}{92.16} & \textcolor{black}{67.14} \\
\bottomrule
\end{tabular*}
\end{table*}

\subsection{Evaluation Metrics}
For PiperX insertion and SO-101 folding, we evaluate the trained models using two metrics. \textcolor{black}{ASR} is the total number of successful attempts divided by the total number of attempts, aggregated across $U_1$--$U_5$ and all trials, \textcolor{black}{reported in \%}. Throughput (TP) is the number of successful tasks per hour.
Each $U_j$ cell represents one individual stage: either a PiperX insertion unit or one SO-101 folding stage. It reports the number of successful completions for that stage and the number of attempts made for that stage. An attempt count of 0 indicates that the trial timed out before reaching this stage.
$N$ is the success target per unit or stage, at which evaluation stops: $N=10$ for PiperX and $N=20$ for SO-101.

\subsection{Data Efficiency of Human Interventions}
\label{sec:exp_scale_intervention}
To evaluate whether targeted human corrections yield greater performance gains than equivalently scaled teleoperated demonstrations under a fixed data-collection budget, we compare two data collection regimes with the same time budget: one collects demonstrations only, while the other reserves part of the budget (an hour) for human corrections. If interventions are more efficient, the same budget should translate into higher policy performance than additional demonstrations alone.
We benchmark a demonstrations-only baseline (SFT) against SFT + Int., which substitutes one hour of teleoperated demonstrations with human-intervention data. The two regimes are evaluated as paired comparisons under identical total budgets, scaled from 2 to 16 hours and trained from the same pretrained checkpoint.
With a four-hour budget, adding one hour of human-intervention data raises ASR from 27.03\% to 51.52\%, and TP from 12.00 to 20.40, compared with the baseline (Table~\ref {tab:scale_intervention}). The eight-hour mixture achieves an ASR of 78.00\%, comparable to 16-hour SFT (78.26\%), while reaching a similar throughput of 53.04 versus 53.27 completed stages per hour.

These gains can be attributed to a fundamental difference between demonstrations and interventions in how they cover the state space. Successful teleoperation may rarely cover states induced by policy errors, whereas interventions supply corrective actions in these states. Their value lies in relevance to the current policy's failures. Since the policy executes autonomously for most of the episode, human supervision intervenes only at a few critical moments, supplying precisely the data the model most needs and thereby increasing the value of the collected training data.

\subsection{Intervention and Real-Robot RL}
\label{sec:exp_complete_loop}
Beyond the data-efficiency comparison, we evaluate how human interventions and reinforcement learning jointly support
continued policy improvement. We conduct ablation experiments on PiperX to assess the contribution of each training
stage (Table~\ref{tab:complete_loop}). All variants start from the same policy trained on demonstrations for 2.5 epochs (SFT).
To evaluate human interventions alone, we fine-tune the policy on demonstrations mixed with one hour of human interventions,
sampling interventions with probability $p=0.5$ (SFT + Int.).
Alternatively, we apply advantage-guided selection to the baseline policy using only demonstration data (SFT + RL): advantages computed from the demonstrations identify \textcolor{black}{high-advantage} segments, which are then sampled with elevated probability during fine-tuning.
The combined setting (Evo-RL) integrates intervention-based fine-tuning with advantage-guided sampling: it mixes original demonstrations, high-advantage demonstration segments sampled with elevated probability, and human-intervention data for policy fine-tuning. \textcolor{black}{High-advantage segments from successful autonomous rollouts are also included} in this round. Finally, we repeat the data collection and training cycle to evaluate the benefits of a second iteration (Evo-RL R2).

\noindent\textit{PiperX insertion.} The ablations in Table~\ref{tab:complete_loop} examine how human corrections and reinforcement learning contribute to policy improvement. Both SFT + Int. and SFT + RL improve over SFT, with human corrections providing the larger gain. Combining the two in Evo-RL yields the highest ASR and TP among the first-round variants, \textcolor{black}{raising ASR from 78.72\% for SFT + Int. to 88.24\% and improving TP by 26.3\%.} These results support complementary roles for corrective experience and advantage-guided refinement: corrections supply behavior for difficulties encountered during deployment, while \textcolor{black}{advantage-guided sampling} directs further training toward promising behavior in the available data.

\textcolor{black}{Beyond these ablations, first-round Evo-RL outperforms RECAP on PiperX, achieving 88.24\% versus 57.78\% ASR and 60.61 versus 35.54 TP. Policy fine-tuning takes approximately 4\,h\,52\,min for Evo-RL (1,511 steps), compared with an estimated 64\,h for RECAP (102,111 steps, 2.5 epochs), both using eight A800 80\,GB GPUs.}

Statistics show that the human-control time fraction decreases from 15.30\% for SFT to 9.56\% for Evo-RL. Together with the higher ASR and TP, this indicates that improved task performance coincides with a smaller share of recorded execution under direct human control. A second collection-and-training round further raises ASR from 88.24\% to 92.16\% and TP from 60.61 to 67.14, with $U_5$ completions increasing from five to eight. Thus, another round can improve execution beyond the first update.

\noindent\textit{SO-101 folding.} The complete Evo-RL method also improves over SFT on folding, raising ASR from 55.56\% to 76.67\% and TP from 75.78 to 90.00. At the spreading stage ($U_2$), successful completions increase from 17 to 19 while attempts decrease from 76 to 36. This substantial reduction in repeated attempts helps explain the higher aggregate ASR and accompanies improved throughput, despite a more modest increase in successful stage completions.

\subsection{Value Learning and Advantage-Guided Sampling}
\label{sec:advantage_selection}
Our value models are task-specific and vision-only. Each model consists of a SigLIP visual encoder followed by an MLP output head, and is trained by randomly sampling frames from demonstrations $\mathcal{D}_{\mathrm{demo}}$, human-intervention episodes $\mathcal{D}_{\mathrm{interv}}$, and autonomous rollouts $\mathcal{D}_{\mathrm{auto}}$. \textcolor{black}{Value training takes approximately 2\,h\,19\,min on three A100 GPUs for Evo-RL, compared with 3\,h\,46\,min on eight A800 80\,GB GPUs for RECAP.}

\textcolor{black}{The} same value \textcolor{black}{model and advantage computation apply to} demonstrations and successful autonomous \textcolor{black}{rollouts.} We \textcolor{black}{jointly} select nonterminal \textcolor{black}{starts with} positive \textcolor{black}{advantages in} the top 10\% across \textcolor{black}{both sources,} merging overlapping or adjacent retained chunks. \textcolor{black}{Selected} autonomous segments \textcolor{black}{join} the original demonstrations and human corrections for \textcolor{black}{fine-tuning;} selected demonstration segments receive \textcolor{black}{increased} sampling \textcolor{black}{weight.}\par
\begin{figure}[!t]
    \centering
    \includegraphics[width=\columnwidth]{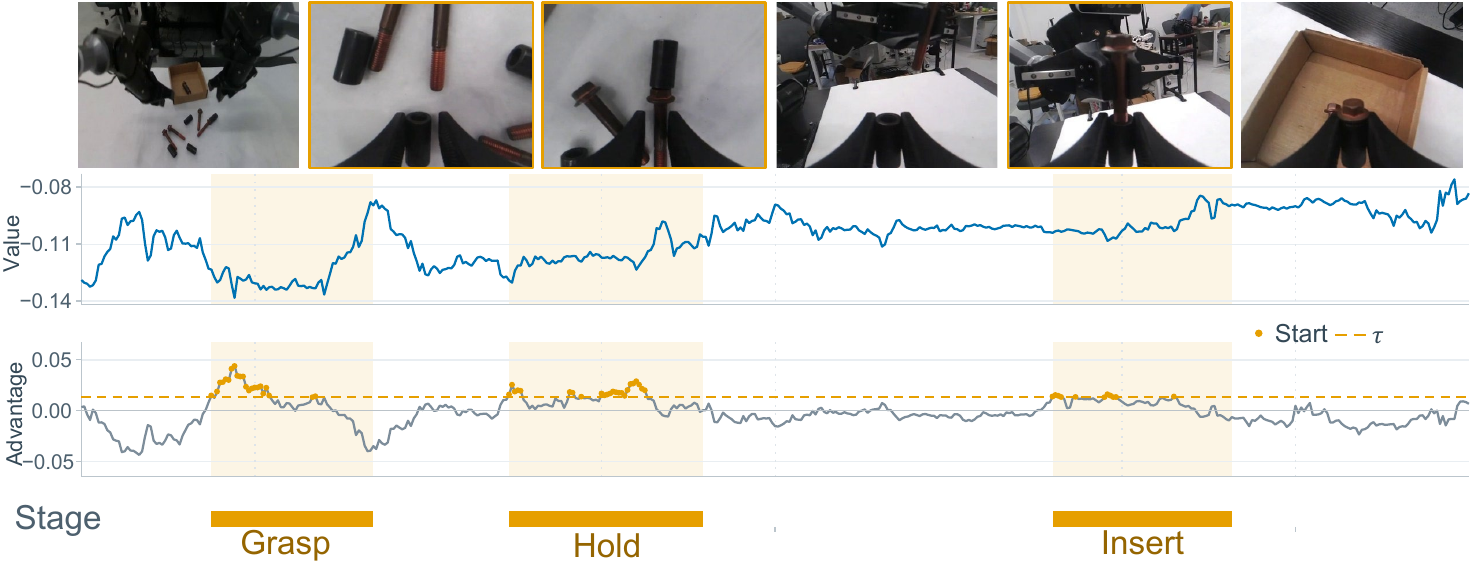}
    \caption{\textbf{Visual value estimation and segment selection.} The visual value function predicts normalized remaining returns to evaluate task states and guide advantage-based training-data selection. Blue and gray curves show value and advantage estimates; gold dots denote selected starts, and the dashed line marks the cutoff $\tau_A$. Gold regions indicate retained segments formed by merging overlapping or adjacent 20-frame forward windows, with bordered images showing corresponding observations.}
    \label{fig:value_segment_selection}
\end{figure}

Figure~\ref{fig:value_segment_selection} plots the value and advantage estimates over a screw-insertion episode. Gold dots mark the selected starts, and the overlaid images show the corresponding observations. Which segments are retained for training is determined by the advantage at each start, not by high state value alone. The selected states consistently correspond to critical transition phases of the task: grasping the target object, manipulating it (inserting the screw into the sleeve), and placing it at the target location. These are precisely the points where small errors most affect downstream execution, and where the policy most needs fine-grained supervision. Notably, states near task completion often exhibit high value yet low advantage, confirming that value alone is insufficient for selecting informative segments.

\section{Conclusion}
We presented Evo-RL, which combines human corrections with advantage-guided experience selection in an iterative policy improvement loop. Corrections address deployment difficulties, while estimated advantages guide subsequent experience selection and sampling. Experiments support their complementary roles: interventions outperform additional demonstrations under matched human-control budgets, and Evo-RL improves attempt success rate and throughput on high-precision and deformable-object manipulation tasks. On PiperX, it outperforms RECAP and either update stage alone, with further gains in a second iteration. To support research on this learning process, RW-RL-Dataset brings together demonstrations, interventions, and autonomous experience, including the training data used in our evaluation.
\bibliographystyle{IEEEtran}
\bibliography{evo_rl_refs}
\end{document}